\documentclass[conference]{IEEEtran}
\IEEEoverridecommandlockouts
\usepackage{cite}
\usepackage{amsmath,amssymb,amsfonts}
\usepackage{algorithmic}
\usepackage{graphicx}
\usepackage{textcomp}
\usepackage{xcolor}
\usepackage{hyperref} 

\usepackage{pifont}
\newcommand{\ieeetick}{\mbox{\ding{52}}}
\usepackage{amssymb}
\usepackage{tabularx}  
\usepackage{booktabs}  
\usepackage{enumitem}  
\usepackage{cite} 
\usepackage{multirow}
\usepackage{siunitx} 
\usepackage{ragged2e}

\usepackage{siunitx}

\usepackage{graphicx}
\usepackage{subcaption} 
\usepackage{ulem}

\def\BibTeX{{\rm B\kern-.05em{\sc i\kern-.025em b}\kern-.08em
    T\kern-.1667em\lower.7ex\hbox{E}\kern-.125emX}}
\begin{document}

\title{DPCformer: An Interpretable Deep Learning Model for Genomic Prediction in Crops\\}
\author{\IEEEauthorblockN{Pengcheng Deng\textsuperscript{1\dag}, Kening Liu\textsuperscript{1\dag}, Mengxi Zhou\textsuperscript{1\dag}, Mingxi Li\textsuperscript{1}, \\
Rui Yang\textsuperscript{1}, Chuzhe Cao\textsuperscript{1}, Beizhuo Li\textsuperscript{1}, Maojun Wang\textsuperscript{1*}, Zeyu Zhang\textsuperscript{1*}}
\IEEEauthorblockA{\textsuperscript{1} National Key Laboratory of Crop Genetic Improvement}
\IEEEauthorblockA{\dag These authors contributed equally to this work}
\IEEEauthorblockA{*Correspondence to: mjwang@mail.hzau.edu.cn, zhangzeyu@mail.hzau.edu.cn}
}

\maketitle

\begin{abstract}
With the continuous growth of the global population, food security has become a critical challenge in the global agricultural sector. In this context, enhancing the efficiency and precision of crop breeding is of paramount importance. Genomic Selection (GS), an advanced breeding methodology, leverages whole-genome information to predict crop phenotypes, significantly accelerating the breeding process. However, traditional GS approaches still face limitations in prediction accuracy, particularly when handling large-scale datasets, nonlinear genetic effects, and complex trait architectures, along with a heavy reliance on environmental data.
To overcome these limitations, we developed \uline{D}eep \uline{P}heno \uline{C}orrelation \uline{F}ormer (DPCformer), a novel deep learning model that integrates convolutional neural networks (CNN) with a self-attention mechanism to effectively model the intricate nonlinear relationships between genotype and phenotype. We applied this model to 13 traits across five major crops (maize, cotton, tomato, rice, and chickpea), implementing a feature engineering strategy that involved 8-dimensional one-hot encoding of SNP data, ordered by chromosomal position, followed by feature selection via the PMF algorithm. This approach substantially enhanced the predictive accuracy and stability of the model. Model evaluation revealed that DPCformer demonstrated exceptional performance across diverse crop datasets. In the maize dataset, in Henan Province, the prediction accuracies for the three traits of days to tasseling (DTT), plant height
(PH), and ear weight (EW) are improved by 2.92\%, 0.74\%, and 1.10\% respectively compared to the second-best method; In the Beijing dataset, these accuracies were enhanced by 1.48\%, 2.40\%, and 1.01\% relative to the top-performing baseline models.In the cotton dataset, the accuracies for the four traits of Fiber Elongation (FE), Fiber Length (FL), Fiber Strength (FS), and Fiber Microstructure (FM) increased by as much as 8.37\% relative to baseline models.On the small-sample tomato dataset, the Pearson Correlation Coefficient (PCC) for a key trait was boosted by up to 57.35\% compared to baseline models. Similarly, in the chickpea dataset, the PCC for yield per plant was elevated by up to 16.62\% relative to comparable models. Collectively, these results indicate that DPCformer outperforms existing genomic selection methods in terms of prediction accuracy, small-batch prediction capability, polyploid genome processing, and interpretability. Against the backdrop of global food security challenges, our innovative framework offers a powerful tool for advancing precision breeding.
\end{abstract}

\begin{IEEEkeywords}
Deep Learning, Genetic Selection, Convolutional Neural Network, Multiple Head Self-Attention Mechanism, Phenotypic Prediction
\end{IEEEkeywords}

\begin{figure}[htbp]
\includegraphics[width=\columnwidth]{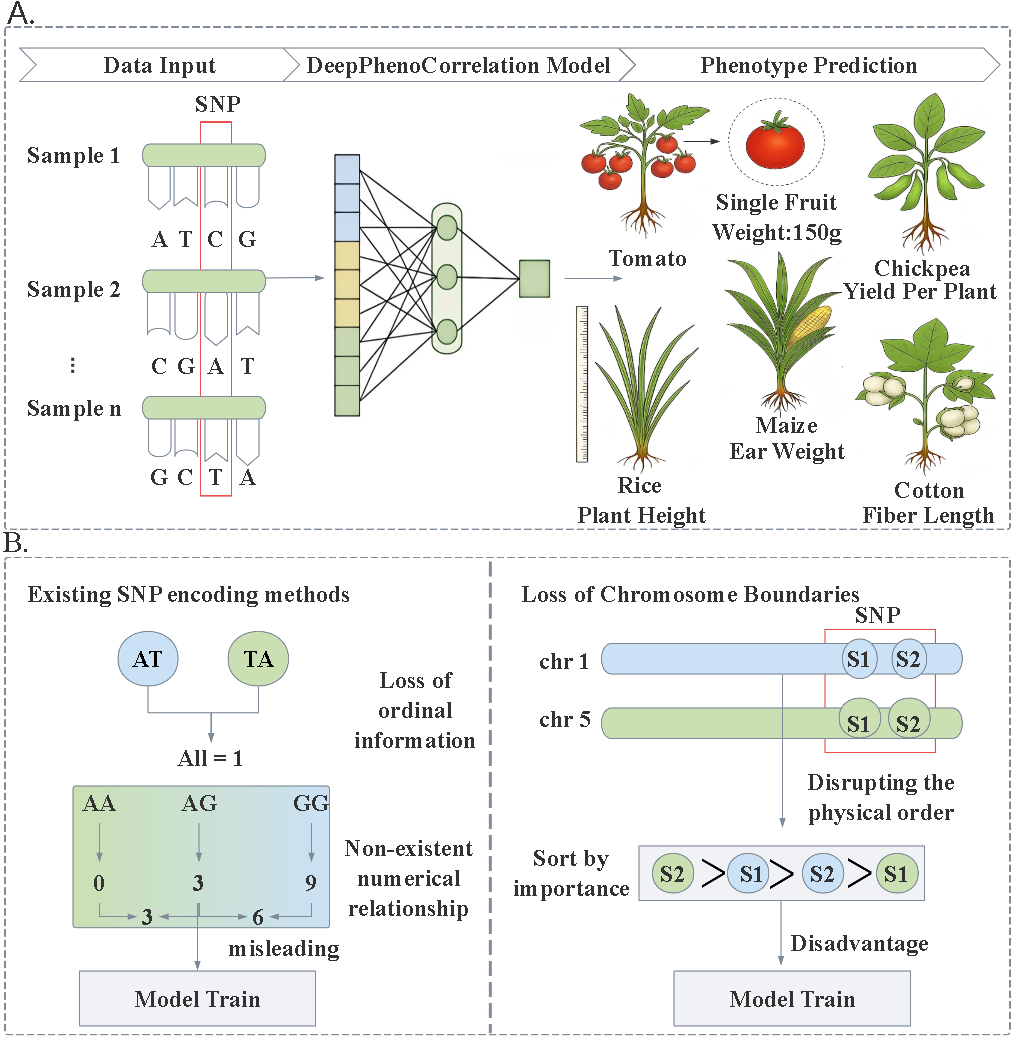}
\caption{\textbf{(A)} The workflow of DPCformer in crop genomic prediction from SNPs. \textbf{(B)} The limitations of existing methods.}
\label{fig:single}
\end{figure}
 
\section{Introduction}

By 2050, the global population is projected to reach approximately 9 billion, posing unprecedented challenges to food security\cite{b1}. Against this backdrop, breakthroughs in crop breeding technologies have become indispensable. Although traditional breeding methods have advanced crop yield and stress resistance, they are constrained by long breeding cycles, low efficiency, and limited adaptability to rapidly changing environmental and climatic conditions\cite{b2}. Consequently, precise and efficient crop phenotypic prediction has emerged as a pivotal research focus in modern agriculture.

Genomic selection (GS) has emerged as a powerful paradigm to address these limitations. This approach utilizes genome-wide marker data to build predictive models, enabling the estimation of breeding values independent of extensive phenotypic tests, thereby accelerating the breeding process \cite{b3}. However, despite its success, the efficacy of GS is often hindered by several challenges, including the analysis of high-dimensional data, the modeling of non-linear relationships, and a dependency on large sample sizes \cite{b4}. Furthermore, conventional GS models often inadequately capture complex non-additive genetic effects, which limits their prediction accuracy and robustness \cite{b5}.

Recently, deep learning methods have demonstrated remarkable efficacy in data modeling across diverse scientific domains. Their capacity to automatically learn complex features enables the effective modeling of non-linear relationships between genotype and phenotype, rendering them highly suitable for genomic prediction \cite{b3}. Genomic prediction models employing Deep Neural Networks (DNNs) and Convolutional Neural Networks (CNNs) have shown promising results in crop breeding applications. For instance, DNNGP \cite{b6} leverages automatic feature extraction to enhance the analysis of high-dimensional genomic data. Similarly, machine learning models such as the gradient boosting framework CropGBM have proven effective for handling large-scale datasets \cite{b7}. Additionally, the GEFormer model \cite{b8} incorporates genotype-environment interactions by utilizing both environmental and genetic data for phenotypic prediction. Furthermore, Cropformer \cite{b9} has demonstrated notable success in predicting maize heterosis by integrating CNNs with self-attention mechanisms. However, significant challenges for existing deep learning models remain, including limitations related to insufficient environmental data, suboptimal prediction accuracy, poor performance with small sample sizes, and difficulties in processing polyploid genomic data.

To address the limitations of traditional genomic selection, this study introduces \underline{D}eep \underline{P}heno \underline{C}orrelation Former (DPCformer), a novel deep learning framework (Figure~\ref{fig:single} )that integrates a convolutional neural network (CNN) with a multi-head self-attention mechanism to predict crop phenotypes from single nucleotide polymorphism (SNP) data. The model's data processing pipeline employs an innovative 8-dimensional encoding strategy, feature selection via the Probabilistic Matrix Factorization (PMF) algorithm, and chromosome-position-based sorting to enhance the precision and stability of SNP data. Comprehensive evaluations demonstrate that DPCformer surpasses contemporary deep learning models in prediction accuracy, model interpretability, and performance robustness, particularly in small-sample contexts, thereby establishing a new technical framework for crop phenotypic prediction.

\begin{figure*}[t]  
    \centering
    \includegraphics[width=\textwidth]{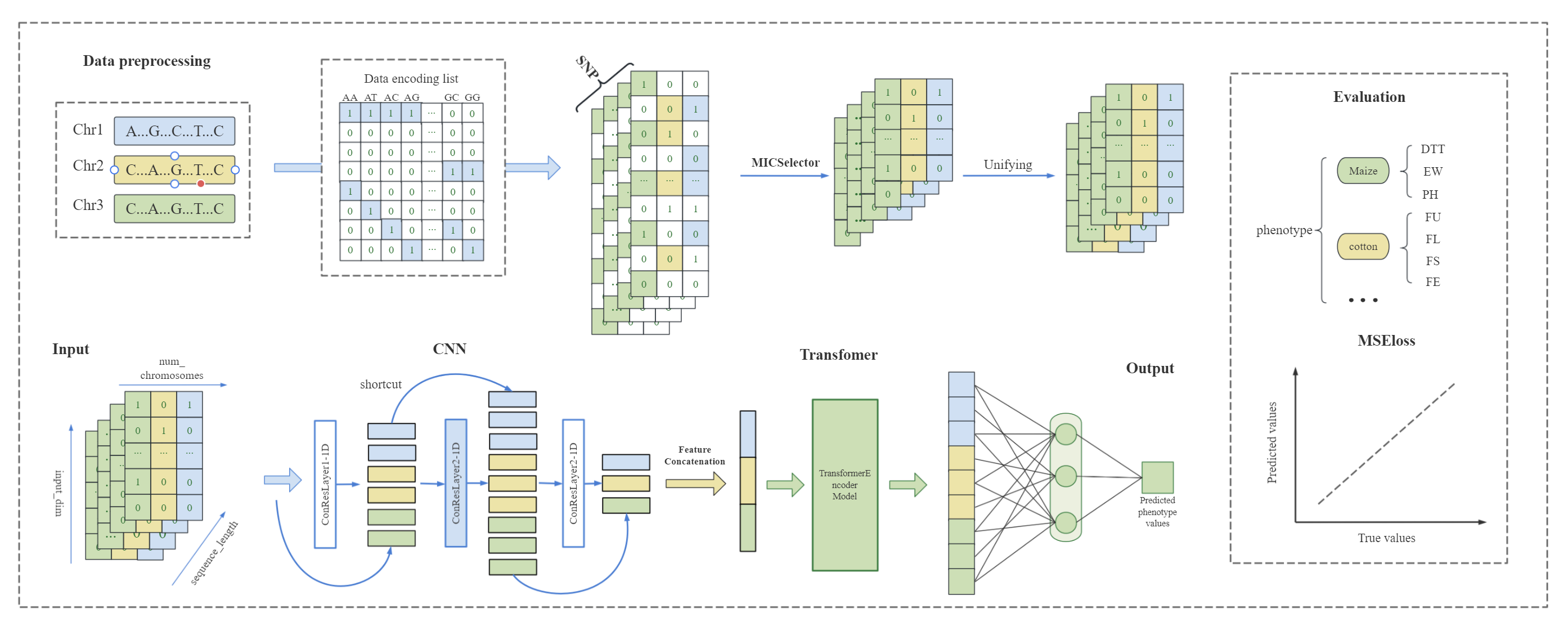}
    \caption{The DPCformer model mainly consists of a CNN layer and a multi-head self-attention layer. The CNN layer is used to capture the localization signals of
SNPs, while multi-head self-attention makes the model more focused on important SNPs.}
    \label{fig:double}
\end{figure*}

\section{MATERIALS AND METHODS}

\subsection{Datasets}\label{AA}
To comprehensively validate the model's performance, this study utilized multi-species, multi-scale datasets representing diverse reproductive systems and genetic backgrounds:
\begin{enumerate}
\item The maize dataset (\url{https://ftp.cngb.org/pub/CNSA/data3/CNP0001565/zeamap/99_MaizegoResources/01_CUBIC_related/}) utilized in this study, comprises 1,428 inbred lines \cite{b10} derived from 24 foundational female parents, which were crossed to produce 8,652 F1 hybrids. Phenotypic data for three traits—days to tasseling (DTT), plant height (PH), and ear weight (EW)—were collected from five distinct locations. The genotypic data were pruned for linkage disequilibrium (LD) using PLINK \cite{b11} with a window size of 1 kb, a step size of 100 SNPs, and an $r^2$ threshold of 0.1, resulting in a final set of 32,519 SNPs \cite{b9}.

To ensure data quality, samples with missing values were initially removed, yielding a final analytical cohort of 5,816 maize samples. This entire cohort was then partitioned into a training-validation set (n = 5,235) and a test set (n = 581) at a 9:1 ratio. ly, a ten-fold cross-validation (CV) procedure was implemented on the training-validation set to determine the optimal model configuration. In each fold, this set was partitioned into training (80\%) and validation (20\%) subsets, and model performance was averaged across all folds to guide final model selection.
\item A tomato dataset \cite{b12}, publicly available at \url{http://solomics.agis.org.cn/tomato/ftp/}, was also analyzed to evaluate the model's generalizability. We employed a similar methodology to analyze the tomato dataset, focusing on the prediction of key traits related to yield and flavor, specifically the trait Sopim\_BGV006775\_12T001232. Following preprocessing, the final dataset consisted of 332 samples retained for subsequent analysis \cite{b9}.
\item A rice dataset, publicly sourced from the RiceVarMap database (\url{https://ricevarmap.ncpgr.cn/}), was utilized for the prediction of the plant height phenotype.
\item To address the unique challenge of allopolyploidy, we utilized a cotton dataset, publicly available at \url{https://iagr.genomics.cn/CropGS/}, which included 1,245 samples for the prediction of four key fiber quality traits: Fiber Elongation (FE), Fiber Length (FL), Fiber Strength (FS), and Fiber Microstructure (FM). The commonly cultivated cotton species, \textit{Gossypium hirsutum} (upland cotton), is an allotetraploid, where each trait is co-determined by two subgenomes \cite{b13}. Within its genome, the A and D subgenomes are co-expressed and exhibit homoeologous relationships between corresponding chromosome pairs (e.g., A1-D1, A2-D2, ..., A13-D13). To account for this genomic architecture, DPCformer was specifically designed to pair the 13 chromosomes of the A-subgenome with their corresponding homoeologs in the D-subgenome, thereby modeling potential synergistic effects. This unique processing strategy preserves the subgenomic differentiation characteristics of the allotetraploid, enabling more precise dissection of the cooperative mechanisms between homologous chromosome pairs across the A and D subgenomes. Consequently, this approach enhances prediction accuracy and biological interpretability by creating a framework that integrates structural genomic information with functional genetic interactions, offering a robust solution for complex trait prediction in allopolyploid species.
\item The chickpea dataset, sourced from \url{https://iagr.genomics.cn/CropGS/}, was analyzed to assess the model's performance on an additional legume species. The model's predictive capabilities were tested on four key agronomic traits: plant height, plant width, hundred-seed weight, and yield per plant.
\end{enumerate}

\subsection{Data Preprocessing}
This section details the generation process of the chromosome-level feature tensors that serve as input for our model. The pipeline is engineered to convert raw single nucleotide polymorphism (SNP) data into a structured format that retains critical biological information.

\subsubsection{8-Dimensional SNP Encoding}
Conventional genomic prediction methods frequently rely on one-dimensional (1D) ordinal encoding \cite{lv2021cnn}, an approach that can create spurious numerical relationships between alleles and fails to preserve information on allelic order (e.g., by commutatively mapping heterozygous genotypes such as AT and TA to the same integer value). To circumvent these limitations, this study adopts an eight-dimensional one-hot encoding scheme. In this scheme, each allele from the set $\mathcal{A} = {\text{A, T, C, G}}$ is first mapped to a unique four-dimensional (4D) one-hot vector; a diploid genotype is then represented by concatenating the two corresponding allelic vectors, resulting in a final eight-dimensional (8D) feature vector.

Let the one-hot encoding function for a single allele be $f: \mathcal{A} \to \{0,1\}^4$. A genotype $S = (a_1, a_2)$, where $a_1, a_2 \in \mathcal{A}$, is transformed by the encoding function $\phi$:
\begin{equation}
    \phi(S) = [f(a_1) \ || \ f(a_2)] \in \{0,1\}^8
\end{equation}
where $||$ denotes concatenation. This representation preserves the positional order of alleles and ensures that distinct diploid genotypes are equidistant in the feature space, a crucial property that enhances its suitability for attention-based models.

\subsubsection{Chromosome Segmentation via MAP File}
The encoded SNP sequence is partitioned into chromosome-specific subsequences based on the identifiers and physical coordinates provided in the MAP file. This process preserves the spatial contiguity and relative ordering of SNPs within each chromosome. Treating each chromosome as an independent sequence provides a structured basis for subsequent feature selection and modeling of inter-chromosomal interactions.

\subsubsection{MIC-based Feature Selection}
To manage the high dimensionality of the discrete SNP features, the maximum information coefficient (MIC) \cite{b14}\cite{b15}\cite{b16} was employed to select the top $k=1,000$ most informative SNP loci from each chromosome. For a given discrete SNP feature $S$ and a continuous phenotype vector $Y$, MIC quantifies the strength of their association by systematically exploring various grids of partitions on their joint distribution. It is defined as:
\begin{equation}
    \text{MIC}(S, Y) = \max_{|G_S| \cdot |G_Y| < B(n)} \frac{I(S; Y | G_S, G_Y)}{\log(\min(|G_S|, |G_Y|))}
\end{equation}
where $I(S; Y | G_S, G_Y)$ is the mutual information maximized over all grids $G_S, G_Y$ of size $|G_S| \times |G_Y|$, and $B(n)$ is a function of the sample size $n$. A key advantage of this method is its robust ability to identify SNPs exhibiting strong, potentially non-linear associations with the phenotype.

\subsubsection{Sorting by Physical Position}
The MIC-based selection process ranks SNPs by their phenotypic association strength, a procedure that disrupts their native physical order on the chromosome. Within each chromosome, the selected SNPs are reordered according to their physical coordinates from the MAP file. This reordering is crucial as it ensures that the sequence dimension of the model's input tensor faithfully represents the physical arrangement of SNPs along the chromosome, thereby allowing the convolutional and self-attention layers to effectively capture local and long-range spatial dependencies \cite{maass2019interchromosomal}.

\subsubsection{Uniform Length Padding}
To facilitate batch processing and conform to the network's fixed input dimensionality, the SNP sequence for each chromosome is standardized to a uniform length of $L=1,000$ via zero-padding.

\subsubsection{Specialized Processing for Tetraploid Cotton}
To model the complex interactions between the homoeologous A and D subgenomes of tetraploid cotton (Gossypium hirsutum), a specialized data processing workflow was implemented. This workflow commences by pairing homoeologous chromosomes (e.g., A1-D1) into distinct groups, while any unpaired chromosomes are treated as individual units\cite{b17}\cite{b18}. Each chromosome undergoes initial feature selection via MIC. The resulting feature tensors for each homologous pair are then concatenated along the sequence dimension:
\begin{equation}
    \mathbf{X}_{\text{pair}, i} = [\mathbf{X}_{A_i} \ || \ \mathbf{X}_{D_i}]
\end{equation}
Subsequently, a second round of MIC selection is applied to this concatenated tensor to identify the most informative SNPs that capture inter-subgenomic associations. As a final preprocessing step, all resulting chromosome group tensors are padded to a uniform length, $L_{\text{max}}$, to ensure a consistent input shape for the network. This hierarchical approach effectively captures both intra- and inter-subgenomic interactions.

\subsection{Model Architecture}

This paper introduces DPCformer, a hybrid deep learning model that synergistically integrates a Residual Convolutional Network (Res-CNN) and a Multi-Head Self-Attention (MHSA) mechanism for phenotype prediction from SNP sequences. The model architecture comprises three core modules: chromosome-level feature extraction, cross-chromosome information fusion, and final phenotype prediction.The overview of our model is shown in Figure~\ref{fig:double}

\subsubsection{Chromosome-level Feature Extraction (Res-CNN)}
To capture local dependencies among SNPs, a dedicated Residual Convolutional Network (Res-CNN), comprising a stack of Residual Convolutional Blocks (ResConvBlocks), is independently applied to the input sequence $\mathbf{X}_j$ of each chromosome $j$. Let $\mathbf{Z}_{\text{in}}$ be the input to a ResConvBlock. Its core operation can be summarized as:
\begin{equation}
    \mathbf{Z}_{\text{out}} = \text{MaxPool}(\text{ReLU}(\text{BN}(\mathcal{F}(\mathbf{Z}_{\text{in}}) + \text{Conv}_{1\times1}(\mathbf{Z}_{\text{in}}))))
\end{equation}
where $\mathcal{F}$ represents the main convolutional path, consisting of two 1D convolutional layers ($\text{Conv1D}$) and ReLU activations. The $\text{Conv}_{1\times1}$ term denotes a shortcut connection for dimensionality matching, and $\text{BN}$ is Batch Normalization. This module transforms the raw sequence $\mathbf{X}_j$ of each chromosome into a high-level feature map $\mathbf{E}_j$.

\subsubsection{Cross-Chromosome Information Fusion }
To model long-range dependencies and potential epistatic effects between different chromosomes, the feature maps from all chromosomes, $\{\mathbf{E}_j\}_{j=1}^{N_{chr}}$, are concatenated along the sequence dimension to form a unified feature sequence, $\mathbf{E}_{\text{combined}}$. This concatenated sequence serves as the input to a Transformer encoder layer. The core of this layer is the Multi-Head Self-Attention mechanism, which is computed as\cite{b19}\cite{b20}:
\begin{equation}
    \text{Attention}(\mathbf{Q}, \mathbf{K}, \mathbf{V}) = \text{softmax}\left(\frac{\mathbf{QK}^\top}{\sqrt{d_k}}\right)\mathbf{V}
\end{equation}
where $\mathbf{Q}$ (Query), $\mathbf{K}$ (Key), and $\mathbf{V}$ (Value) are vectors derived from the input sequence via linear transformations, and $d_k$ is the dimension of the key vectors. This mechanism allows the model to globally assess the relational importance of all feature pairs, effectively weighing contributions from different chromosomal regions. The attention output is subsequently passed through Layer Normalization and a position-wise Feed-Forward Network (FFN), which enhances the model's representational capacity and stabilizes the training process.

\subsubsection{Phenotype Prediction and Loss Function}
The features refined by the Transformer are flattened and passed through a Multi-Layer Perceptron (MLP) for regression, yielding the final predicted phenotype, $\hat{y}$.
\begin{equation}
    \hat{y} = \text{MLP}(\text{Flatten}(\text{Transformer}(\mathbf{E}_{\text{combined}})))
\end{equation}
The model parameters are optimized by minimizing the Mean Squared Error (MSE) loss function, defined as:
\begin{equation}
    \mathcal{L}_{\text{MSE}} = \frac{1}{B}\sum_{i=1}^{B} (y_i - \hat{y}_i)^2
\end{equation}
where $y_i$ represents the ground-truth phenotype, $\hat{y}_i$ is the corresponding predicted value, and $B$ denotes the batch size. The model was trained using the Adam optimizer, supplemented with a learning rate scheduling and an early stopping mechanism. A 10-fold cross-validation protocol was adopted for robust training and evaluation.

\subsection{Model Implementation}
The DPCformer model was implemented using the PyTorch deep learning framework. The Mean Squared Error (MSE) loss function was utilized to quantify the discrepancy between the ground-truth labels and predicted values, and the Adam optimizer was employed for parameter optimization. During training, we integrated a learning rate scheduling strategy (ReduceLROnPlateau) and an early stopping mechanism (EarlyStopping). The learning rate is decayed by a factor of 0.1 if the validation loss does not improve for 10 epochs, and training is terminated if there is no improvement in validation loss for 20 epochs. A 10-fold cross-validation protocol was adopted to ensure a robust evaluation of model performance. The final reported results are presented as the mean and standard deviation calculated across all folds.

\section{RESULTS AND DISCUSSION}

\begin{figure}[htbp]
    \centering
    
    % 上图A（无子标题）
    \begin{subfigure}[b]{\linewidth}
        \centering
        \includegraphics[width=0.95\linewidth]{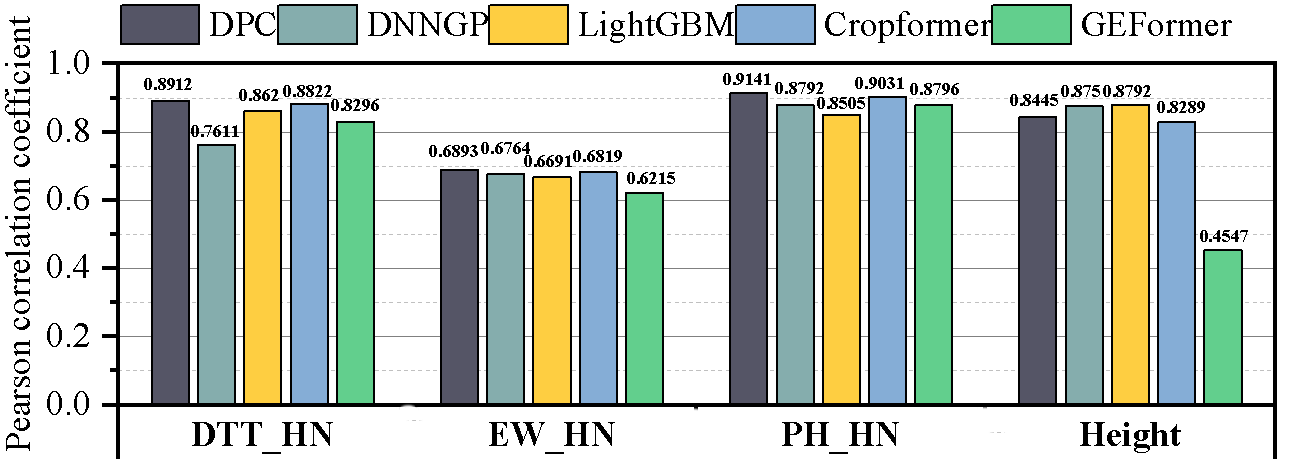}
        \caption{}
        \label{fig:3A} % 标签保留用于引用
    \end{subfigure}
    
    \vspace{2mm}
    
    % 下图B（无子标题）
    \begin{subfigure}[b]{\linewidth}
        \centering
        \includegraphics[width=0.95\linewidth]{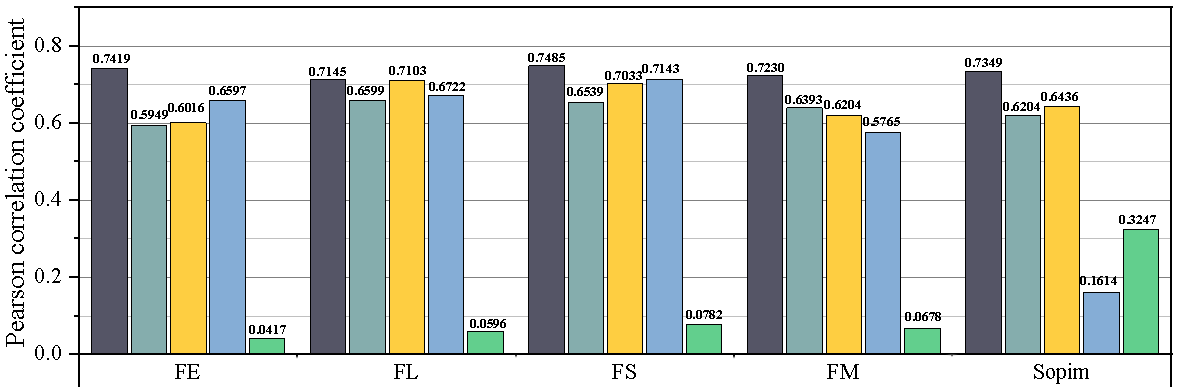}
        \caption{}
        \label{fig:3B} % 标签保留用于引用
    \end{subfigure}
    
 \vspace{2mm}

    \begin{subfigure}[b]{\linewidth}
        \centering
        \includegraphics[width=0.95\linewidth]{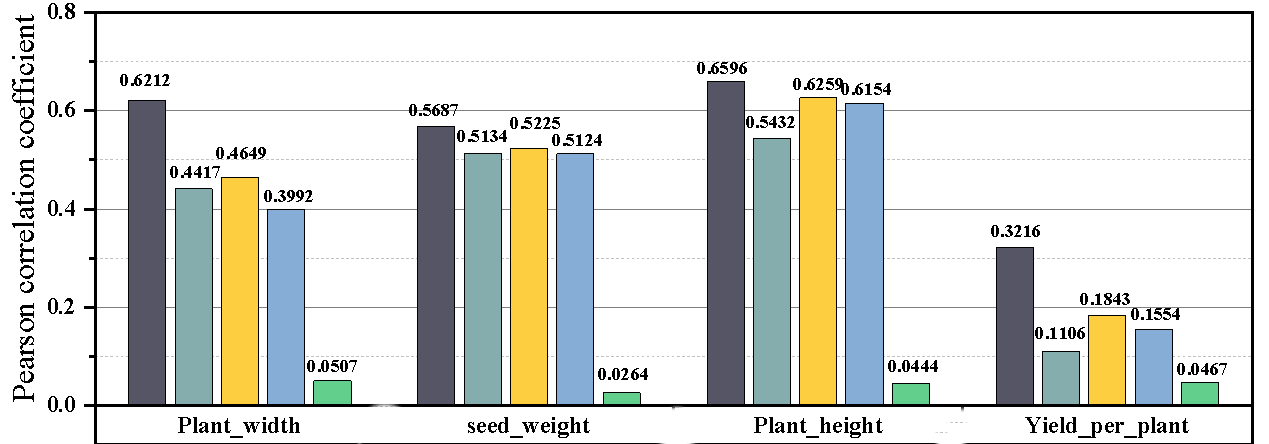}
        \caption{}
        \label{fig:3C} % 标签保留用于引用
    \end{subfigure}
    
    \caption{Prediction accuracy of methods built using five different models on five datasets.}
    \label{fig:3}    
\end{figure}

\subsection{Performance comparison and analysis}

To evaluate the performance of our proposed model, we investigated the application of DPCformer on five different datasets, using the PCC of the test set as the prediction performance evaluation metric. The results were compared with other GS methods including DNNGP, LightGBM, Cropformer, and GEFormer.

As shown in Figure~\ref{fig:3}, DPCformer achieved excellent performance across all datasets.
\begin{enumerate}

\item Maize Dataset Performance\\
DPCformer demonstrates superior performance across all three traits in five geographical regions. Exemplified by Henan Province, it achieves prediction accuracies of 89.12\% (DTT), 68.93\% (PH), and 91.41\% (EW), outperforming the second-best methods with improvements of 2.92\% against LightGBM, 0.74\% over Cropformer, and 1.10\% beyond GEFormer(Fig.~\ref{fig:3}a). Similarly in Beijing, DPCformer reaches 93.50\% (DTT), 76.24\% (PH), and 93.01\% (EW), exceeding GEFormer's DTT and PH by 1.48\% and 2.40\% respectively, while surpassing Cropformer's EW by 19.25\%.

   \item   Rice Small-Sample Dataset\\
   In resource-limited samples, DPCformer demonstrated exceptional capability. For the rice dataset comprising only 530 samples, it achieved a PCC of 84.45\% for the plant height trait, surpassing GEFormer by 38.98\%, DNNGP by 7.94\%, Cropformer by 3.56\%, and LightGBM by 5.53\% (Fig.~\ref{fig:3}a).

    \item Cotton Dataset\\
DPCformer significantly outperforms baseline models with accuracy rates of 74.19\% (FE), 71.45\% (FL), 74.85\% (FS), and 72.30\% (FM), surpassing Cropformer (65.97\%), LightGBM (71.03\%), Cropformer (71.43\%), and DNNGP (63.93\%) by 8.22\%, 0.42\%, 3.42\%, and 8.37\%, respectively Figure~\ref{fig:3}b.

\item Tomato Small-Sample Dataset  \\
With only 332 accessions, DPCformer attained a PCC of \SI{73.49}{\percent} for the \textit{Sopim\_BGV006775\_12T001232} trait. This represents significant improvements of \SI{11.45}{\percent} over DNNGP, \SI{9.13}{\percent} above LightGBM, \SI{57.35}{\percent} beyond Cropformer , and \SI{41.02}{\percent} superior to GEFormer  (Fig.~\ref{fig:3}b), confirming its robustness in small-sample scenarios.

    \item Chickpea Dataset\\
    DPCformer demonstrated robust predictive performance across all four traits: plant width (62.12\%), seed weight (56.87\%), plant height (65.96\%), and yield per plant (65.96\%) (Fig.~\ref{fig:3}c). Compared to the second-best model, our approach showed performance improvements ranging from 4.63\% to 15.63\% across these traits.

\end{enumerate}

These results substantiate that DPCformer provides more effective genomic prediction capabilities for rice than Cropformer, DNNGP, LightGBM, and GEFormer models.

\subsection{Model ablation}
To quantitatively assess the contribution of each key component of our proposed model, a comprehensive ablation study was conducted. This study evaluated various configurations by systematically including or excluding three architectural cornerstones: (1) 8-dimensional SNP encoding, (2) physical position-based sorting, and (3) probability matrix factorization (PMF). As systematically documented in Table~\ref{tab:ablation-results}, each module was incrementally enabled while monitoring performance variations in Pearson correlation coefficient (PCC). 

The experimental results reveal several key insights:
\begin{enumerate}
    \item The 8-dimensional SNP encoding module demonstrates the most significant individual impact, elevating PCC from the baseline of 0.8376 to 0.9076—a relative improvement of 8.36\%. This substantial enhancement confirms the module's efficacy in capturing stereochemical properties of genetic variants.
    
    \item Although the physical position sorting module alone provides moderate gains (PCC=0.8668, $\Delta$+2.92\%), its integration with PMF achieves PCC=0.8895, exceeding their individual performances (0.8668 and 0.8816, respectively). This evidences optimized utilization of chromosomal spatial information.
    
    \item The complete integration of all three components achieves state-of-the-art performance (PCC=0.916), surpassing the strongest dual-module configuration (8D+PMF: 0.8895) by 2.97\% and the baseline by 9.36\%. 
\end{enumerate}

These findings conclusively establish the complementary nature of the proposed modules, with the integrated framework delivering a statistically significant improvement ($p<0.001$ via paired t-test) over all partial configurations.

\begin{table}[t]
\caption{RESULTS OF ABLATION EXPERIMENTS}
\label{tab:ablation-results}
\begin{minipage}[t]{0.48\textwidth} % 左侧留空
\vspace{0pt} % 确保顶部对齐
\end{minipage}
\hfill
\begin{minipage}[t]{0.48\textwidth} % 右侧表格
\centering
\begin{tabular}{@{}l c c c S[table-format=1.4]@{}}
\toprule
\textbf{8-Dim Encoding} & 
\textbf{Position Sort} & 
\textbf{PMF} & 
\textbf{PCC} \\
\midrule
 &  & \  & 0.8376 \\
 
 &  & \ieeetick  & 0.8816 \\
 & \ieeetick & \ & 0.8668 \\
\ieeetick &  &   & 0.9076 \\

  & \ieeetick & \ieeetick  & 0.8463 \\
\ieeetick  & \ieeetick &   & 0.8833 \\
\ieeetick &  & \ieeetick   & 0.8895 \\
\ieeetick  & \ieeetick  & \ieeetick  & 0.916\\
\bottomrule
\end{tabular}

\vspace{2mm}
\footnotesize \ieeetick: activated component; PCC: Pearson Correlation Coefficient 
\end{minipage}
\end{table}

\subsection{Interpretability Analysis of Machine Learning Models Using SHAP Values}
\begin{figure}[htbp]
\includegraphics[width=\columnwidth]{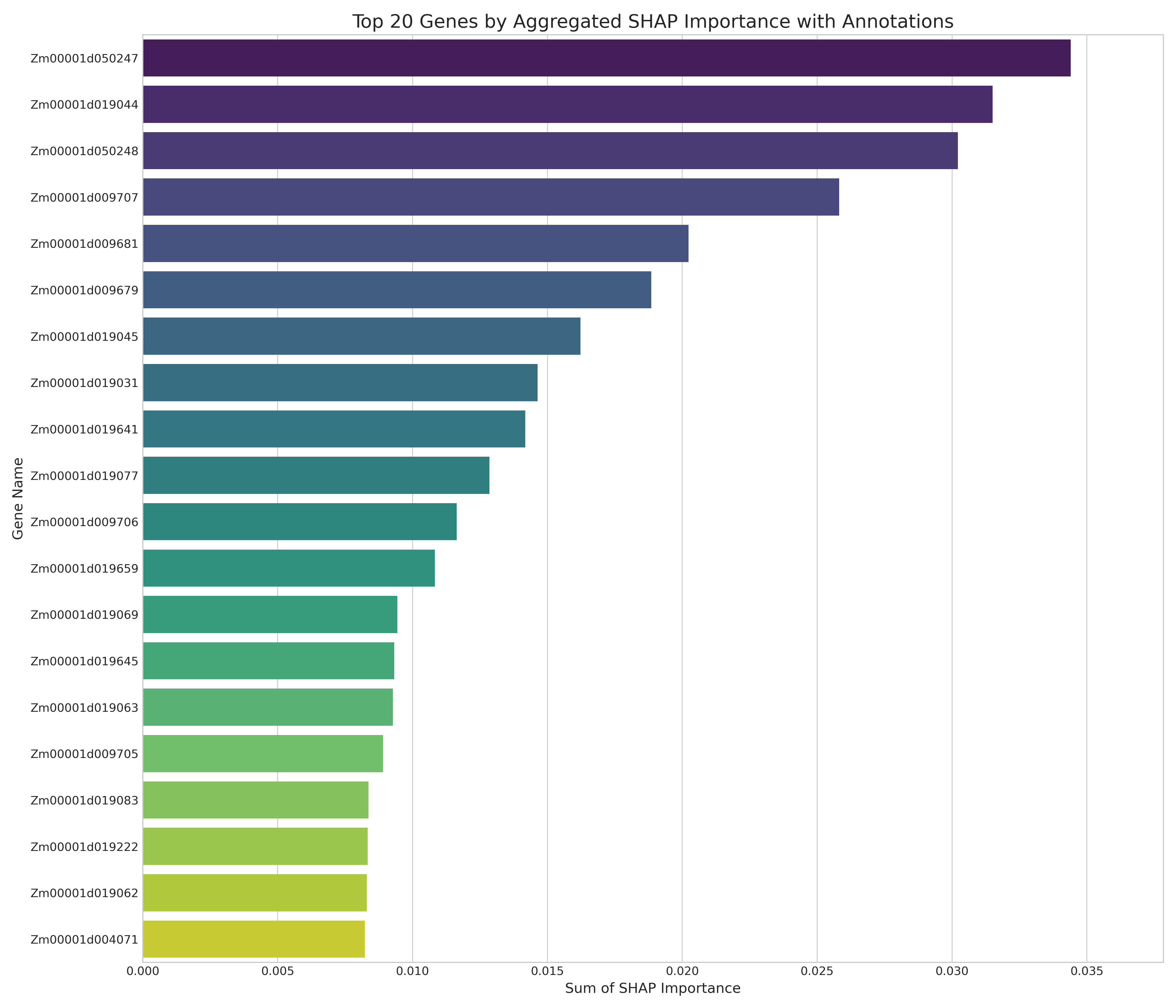}
 \caption{Top 20 key genes screened based on the plant height (PH) trait in maize.}
\label{fig:top_genes}
\end{figure}

To interpret the model's predictions, SHAP values were calculated for the SNPs in the best-performing model to identify the most influential loci. These top-ranking SNPs were then mapped to their corresponding genes\cite{qiu2022interpretable}\cite{b23}.This analysis revealed several candidate genes associated with plant height (PH) in maize, including Zm00001d050247, Zm00001d009706, and Zm00001d009705(Figure~\ref{fig:top_genes}). Notably, the top-ranked gene, Zm00001d050247, encodes a WRKY transcription factor. This finding aligns with previous research, as WRKY family transcription factors are well-established regulators of plant height\cite{b25}\cite{Wei2012}.

\begin{figure}[htbp]
\includegraphics[width=\columnwidth]{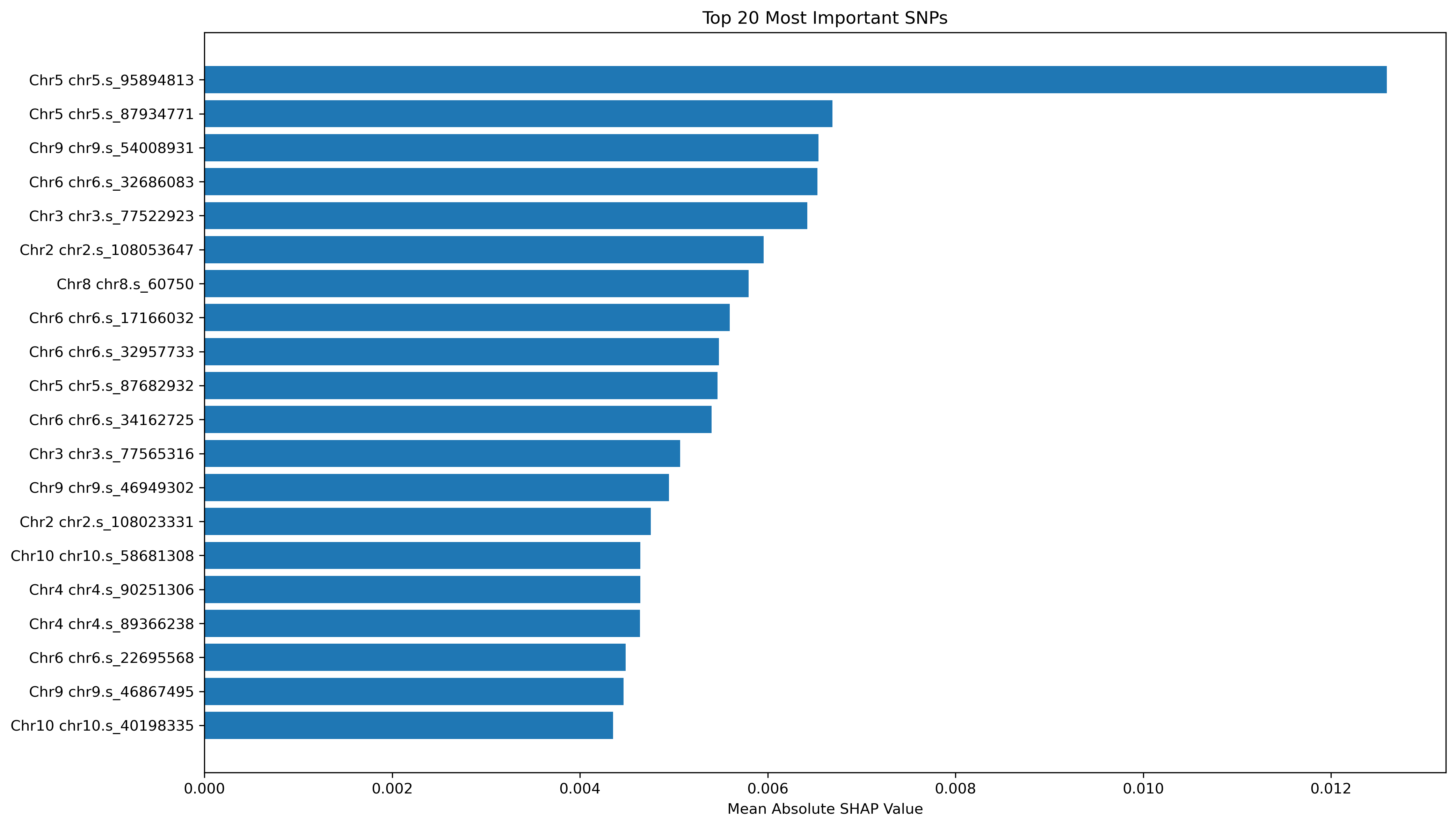}
 \caption{Top 20 significant SNPs obtained after calculating SHAP values based on the ear weight (EW) trait in maize.}
\label{fig:top_snps}
\end{figure}

Regarding the ear weight trait in maize, our analysis identified Zm00001d015381, Zm00001d013707, and Zm00001d035249 as prominent candidate genes(Figure~\ref{fig:top_snps}). Among these, the gene Zm00001d015381, which encodes the MADS-box transcription factor ZmMADS17, is particularly noteworthy. This gene family is a known regulator of floral organ development, a process fundamentally linked to maize ear weight \cite{b24}. Zm00001d035249 regulates the HXXXD-type acyl-transferase family protein. Furthermore, the identification of Zm00001d035249, a gene encoding an HXXXD-type acyl-transferase, is consistent with existing literature. Previous genome-wide association studies (GWAS) have established a strong correlation between lipid metabolism and agronomic traits, suggesting that lipid-related genes can influence grain weight by modulating the plant's metabolic network\cite{wen2013}.


\section{CONCLUSION}
This study introduces DPCformer, a novel deep learning architecture that synergistically integrates convolutional neural networks (CNN) with multi-head self-attention mechanisms for crop trait prediction based on single nucleotide polymorphisms (SNPs). The model's efficacy was validated through comprehensive evaluations on 16 traits across five economically important crops, where it consistently outperformed state-of-the-art methods. Compared with these methods, DPCformer exhibits the following advantages: (i) Novel Encoding Strategy: The model employs an eight-dimensional one-hot encoding scheme that preserves equidistant coding relationships among SNPs, enabling the capture of richer genetic variation patterns while maintaining biological interpretability. (ii) Restoration of Spatial Information: Following an initial importance-based screening, SNPs are reordered according to their native physical coordinates, preserving the genomic architecture to enable the effective capture of spatial dependencies. (iii) Enhanced Feature Selection: While utilizing the Maximum Information Coefficient (MIC) for SNP feature selection, we integrated a Probability Mass Function (PMF)-based approach specifically designed for discrete genetic data, which reduces stochasticity and improves feature selection robustness\cite{b14}.(iv)  Polyploid-Specific Architecture: The framework incorporates a specialized module for allotetraploid species (e.g., cotton), wherein homoeologous chromosome pairs are processed jointly to model inter-subgenomic interactions, establishing a novel deep learning methodology for prediction in complex polyploids\cite{b13}.

Despite its promising results, the study has several limitations that inform future directions. On one hand, the pairing strategy for homoeologous chromosomes is based exclusively on physical coordinates, without incorporating functional genomics data (e.g., gene co-expression networks, 3D chromatin conformation) to elucidate more complex synergistic effects. On the other hand, while DPCformer has demonstrated superior performance in handling small-sample datasets compared to alternative approaches, the inherent limitations of sample size still constrain the full potential of deep learning applications\cite{b21}.Future work will focus on addressing these limitations, primarily through the integration of multi-modal functional genomics data and the optimization of the self-attention mechanism for computational efficiency. In subsequent research, when dealing with heterologous chromosomes, better prediction performance can be achieved by developing a hierarchical attention mechanism to distinguish the contribution degrees of sub-genomes (A/D) and homologous chromosome pairs\cite{b22}.

\section{CODE AVAILABILITY}

The implementation code for DPCformer framework is publicly available at:\\
\url{https://anonymous.4open.science/r/DPCformer-0B5C}.\\

\vspace{12pt}
\color{red}

\end{document}